\documentclass[10pt,twocolumn,letterpaper]{article}

\usepackage[pagenumbers]{cvpr} 

\usepackage{color}

\definecolor{cvprblue}{rgb}{0.21,0.49,0.74}
\usepackage[pagebackref,breaklinks,colorlinks,allcolors=cvprblue]{hyperref}
\usepackage[marginal]{footmisc}
\def\paperID{4697} 
\def\confName{CVPR}
\def\confYear{2025}

\title{1-2-1: Renaissance of Single-Network Paradigm for Virtual Try-On} 

\author{
    Shuliang Ning\textsuperscript{1,2}   \qquad
    Yipeng Qin\textsuperscript{3} \qquad 
    Xiaoguang Han\textsuperscript{2,1,\footnotemark[1]} \\
\textsuperscript{1} FNii, CUHKSZ \qquad
\textsuperscript{2} SSE, CUHKSZ \qquad
\textsuperscript{3} Cardiff University \qquad
}

\begin{document}

\twocolumn[{
\renewcommand\twocolumn[1][]{#1}
\maketitle
\begin{center}
    \captionsetup{type=figure} 
    \includegraphics[width=.955\textwidth]{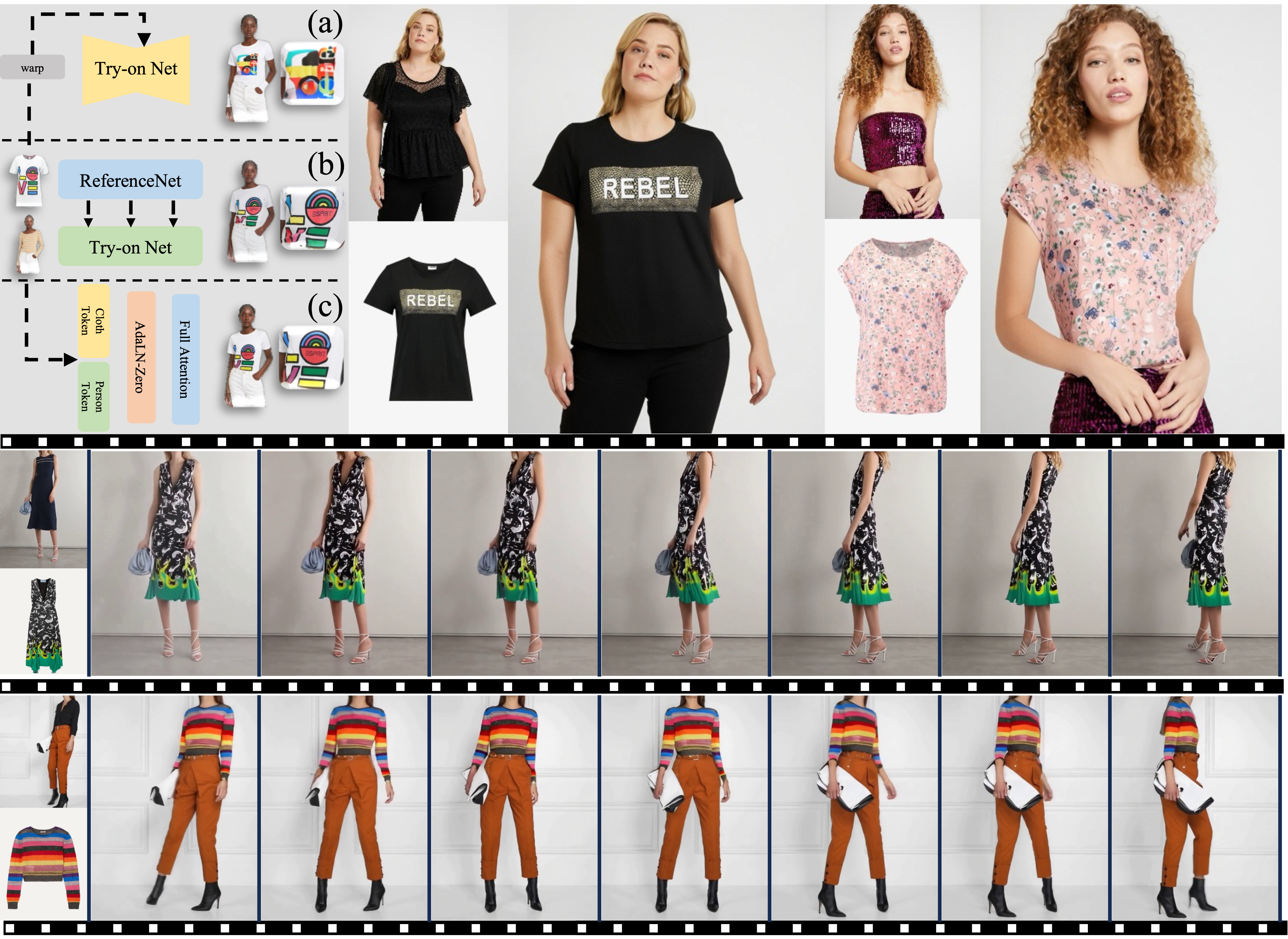}
    \vspace{-2mm}
    \caption{Examples of image and video virtual try-on (VTON) results on VITONHD \cite{choi2021viton}  and VIVID \cite{fang2024vivid} datasets. (a): Warping-based Try-On; (b): ReferenceNet-based Try-On; (c) MN-VTON (Ours). Our MN-VTON marks a renaissance of the {\it single-network paradigm} for VTON used in (a) but discarded in (b).
    The rest of the figure shows the results of our MN-VTON for both image and video VTON.
    }
    \label{fig:teaser}
    \vspace{-2mm}
\end{center}
}]

\footnotetext[1]{Corresponding Author.}

\begin{abstract}

Virtual Try-On (VTON) has become a crucial tool in e-commerce, enabling the realistic simulation of garments on individuals while preserving their original appearance and pose. Early VTON methods relied on {\bf single} generative networks, but challenges remain in preserving fine-grained garment details due to limitations in feature extraction and fusion. 
To address these issues, recent approaches have adopted a {\bf dual}-network paradigm, incorporating a complementary ``ReferenceNet'' to enhance garment feature extraction and fusion. While effective, this dual-network approach introduces significant computational overhead, limiting its scalability for high-resolution and long-duration image/video VTON applications.
In this paper, we challenge the dual-network paradigm by proposing a novel {\bf single}-network VTON method that overcomes the limitations of existing techniques. Our method, namely MN-VTON, introduces a Modality-specific Normalization strategy that separately processes text, image and video inputs, enabling them to share the same attention layers in a VTON network. 
Extensive experimental results demonstrate the effectiveness of our approach, showing that it consistently achieves higher-quality, more detailed results for both image and video VTON tasks. Our results suggest that the single-network paradigm can rival the performance of dual-network approaches, offering a more efficient alternative for high-quality, scalable VTON applications.


\end{abstract}   
\section{Introduction}

\label{sec:intro}


Virtual Try-On (VTON) enables the seamless alteration of a person's clothing in an image or video, allowing different garments to be realistically simulated while preserving the individual's original appearance and pose~\cite{han2017viton, lee2022hrviton}. 
As e-commerce expands and demand for personalized shopping grows, VTON has become a vital tool in online retail. For merchants, it enables showcasing multiple garments on the same model without costly photoshoots, simplifying inventory and enhancing visual appeal. Meanwhile, AI-driven VTON also appeals to casual users, allowing them to explore styles interactively and create engaging content.

Early methods~\cite{han2017viton, wang2018toward, choi2021viton, xie2021towards, lee2022hrviton, xie2023gpvton, zhao2021m3d} laid the groundwork for image-based VTON by employing a {\bf single} generative network to synthesize realistic results. 
These methods leverage the capabilities of deep generative models, such as GANs~\cite{goodfellow2014generative}, by first warping the garment to align with the target person’s pose, and then blending the warped garment onto the target person’s image using a generative network. 
However, recognizing that the features of warped garment images can be ``washed away'' during network propagation~\cite{park2019semantic}, researchers proposed layer-wise conditioning techniques that incorporate additional garment features throughout the generation process, significantly advancing VTON results~\cite{gou2023taming, morelli2023ladi}. 
For instance, LaDI-VTON~\cite{morelli2023ladi} utilizes a CLIP~\cite{radford2021learning} encoder to extract features from both the garment image and its textual description, using these as conditioning inputs to enhance realism in the generated results.
Despite these advancements, single-network methods still face challenges in preserving fine-grained garment textures, largely due to three inherent limitations: 
\begin{itemize}
    \item {\bf [L1]} CLIP is designed for high-level semantic representation, often neglecting the subtle textural details crucial for garment realism. 
    \item {\bf [L2]} Applying the same generalized CLIP features uniformly across all layers fails to account for the unique characteristics of each layer, resulting in a mismatch.
    \item {\bf [L3]} CLIP handles text and image inputs similarly, neglecting the intrinsic differences between the visual and textual modalities, resulting in suboptimal fusion and texture preservation.
\end{itemize} 

State-of-the-art methods~\cite{zhu2023tryondiffusion, kim2024stableviton, xu2024ootdiffusion, choi2024improving, zhang2024mmtryon, zhu2024m} address these three limitations with a {\bf dual} networks paradigm, which introduces a complementary {\it ReferenceNet} with the same architecture as the main generative network.
Specifically, this ReferenceNet addresses {\bf [L1]} by being trainable, allowing it to learn and extract the textual details crucial for VTON; {\bf [L2]} by extracting multi-level features from the garment image and feeding them into the corresponding layers of the main generative network, respectively; and {\bf [L3]} by using only the garment image as input, which ensures a single-modality, focused feature extraction process.
This paradigm has also been adopted by state-of-the-art video VTON methods~\cite{xu2024tunnel, zheng2024viton, fang2024vivid}, which extend image-based VTON approaches by incorporating temporal attention modules to ensure continuity and coherence across video sequences.
To date, the {\it dual-network paradigm} to extract garment features has become a \textbf{common belief} in VTON. 
However, while effective, the inclusion of an additional ReferenceNet introduces substantial computational overhead. 
This becomes an increasingly significant concern as user demands for {higher-resolution}, {greater-fidelity}, and {longer-duration} images and videos continue to grow, posing a major challenge for the practical deployment of such methods in real-world scenarios.

In this paper, we challenge the above-mentioned belief by proposing a novel method, namely MN-VTON, which achieves high-quality VTON with a \textbf{single} generative network, marking a renaissance of {\it single-network paradigm} for VTON. 
Specifically, we address {\bf [L3]} by introducing a novel Modality-specific Normalization strategy that normalizes and modulates different input modalities (text, image/video) separately, enabling them to be effectively processed by the {\it same} attention layer; {\bf [L2]} by nature as the features extracted at each layer correspond to its specific characteristics; and {\bf [L1]} by nature as the network layers are trainable, allowing them to learn and extract the textual details crucial for VTON.
Extensive experiments on two image datasets (VITONHD~\cite{choi2021viton}, DressCode~\cite{morelli2022dresscode}) and two video datasets (VVT~\cite{wu2022fw}, VIVID~\cite{fang2024vivid}) demonstrate the effectiveness of our method.
Our contributions include:
\begin{itemize}
\item We challenge the {\it dual network paradigm} belief in VTON and propose a novel method that achieves high-quality VTON with a {\it single} network, marking a renaissance of {\it single-network paradigm} for VTON (\ie, 1-2-1).
\item We propose a novel modality-specific normalization strategy that normalizes and modulates different input modalities (text, image/video) separately, enabling them to be effectively processed by the {\it same} attention layers.
\item Extensive experiments demonstrate that our method consistently delivers higher quality and more detailed results for image VTON tasks, as well as higher fidelity and higher resolution for video VTON tasks.
\end{itemize}

\section{Related Work}

\noindent\textbf{Single-Network Virtual Try-On.}
Image-based virtual try-on (VTON) methods aim to generate realistic images of a person wearing a specified garment while preserving the individual’s original pose and identity. Early approaches established the foundation for VTON by utilizing a {\bf single} generative network to synthesize lifelike try-on results. These early models employed techniques ranging from feed-forward warping networks \cite{han2017viton, wang2018characteristicpreservingimagebasedvirtualtryon,ge2021parserfreevirtualtryondistilling, xie2023gpvtongeneralpurposevirtual} to diffusion-based iterative sampling \cite{zhang2023warpdiffusionefficientdiffusionmodel,zhu2023tryondiffusion,morelli2023ladivtonlatentdiffusiontextualinversion,kim2023stablevitonlearningsemanticcorrespondence,zhang2024mmtryonmultimodalmultireferencecontrol,xu2024ootdiffusionoutfittingfusionbased, chong2024catvtonconcatenationneedvirtual}, each contributing essential advancements toward realistic and reliable VTON experiences.
Specifically, warping-based methods are typically divided into two stages: a clothing warping network that adapts the garment to fit the person's body shape, and a generative try-on network that seamlessly blends the warped clothing with the human model. Notably, VITON \cite{han2017viton} first introduced the use of the Thin Plate Spline (TPS) transformation to effectively deform clothing to fit the target body. 
Building on this, CP-VTON \cite{wang2018characteristicpreservingimagebasedvirtualtryon} refined the VTON process by explicitly dividing it into warping and generation stages, enhancing control over garment fit and appearance. 
More recently, GP-VTON \cite{xie2023gpvton} proposed the Local-Flow Global-Parsing (LFGP) warping module, achieving semantically accurate garment deformation even with challenging inputs.
With the advent of diffusion models, researchers have explored these as replacements for traditional feed-forward networks as the try-on module. Both \cite{morelli2023ladi, gou2023taming} integrate warping modules with diffusion-based try-on networks, producing high-quality, realistic try-on results that significantly enhance visual fidelity.
In addition, \cite{morelli2023ladi} recognizes that the features of warped garment images can be ``washed away'' during network propagation~\cite{park2019semantic} and proposed a CLIP-based layer-wise conditioning technique that incorporates additional garment features throughout the generation process to improve VTON results.

Despite these advancements, single-network methods face three key limitations (\textbf{[L1]}, \textbf{[L2]}, and \textbf{[L3]}) as outlined in Sec.~\ref{sec:intro}, which have driven the development of the dual-network paradigm for VTON.

\vspace{1mm}
\noindent\textbf{Dual-Network Virtual Try-On.}
TryOnDiffusion \cite{zhu2023tryondiffusion} first introduces a garment encoder that extracts clothing features and employs cross-attention mechanisms to merge the garment and person features, enabling implicit warping and try-on of clothes. Building on \cite{zhu2023tryondiffusion} and \cite{hu2024animate}, recent methods \cite{choi2024improving, kim2024stableviton, xu2024ootdiffusion, zhang2024mmtryon} leverage a {\bf dual}-network paradigm, which either fully or partially replicates the original U-Net structure. 
Experiments with these approaches demonstrate that the use of well-aligned, multi-scale features significantly improves the detail and realism of generated clothing.
In addition, this dual-network paradigm has proven effective not only for image-based VTON but also for video VTON, where it is further enhanced by incorporating a temporal attention module to capture and model temporal shifts in the video.

However, the inclusion of an additional network introduces significant computational overhead. As the demand for higher-resolution, higher-fidelity, and longer-duration images and videos continues to grow, the limitations of dual-network methods have become increasingly apparent.

\vspace{1mm}
\noindent\textbf{Video Generation and Virtual Try-On.}
The marriage of video generation and video virtual try-on (VTON) begins with a paradigm where the initial image is generated by a try-on network~\cite{sun2024outfitanyoneultrahighqualityvirtual}, followed by the generation of the full video using pose-guided video diffusion models~\cite{hu2024animateanyoneconsistentcontrollable}. This approach primarily emphasizes maintaining body consistency across frames, with less focus on garment details. Subsequent works~\cite{fang2024vivid, xu2024tunnel} have enhanced this paradigm by incorporating the try-on functionality directly into pose-guided video generation, creating end-to-end systems. While these methods show promising results, they may struggle with vigorous human motion and complex garment textures. 
Despite employing a temporal module across the time dimension, it often functions as a local attention mechanism within the spatial region.
Recent advancements in text-to-video generation~\cite{yang2024cogvideox, opensora} highlight the effectiveness of DiT~\cite{peebles2023scalablediffusionmodelstransformers} and full attention mechanisms across both temporal and spatial dimensions. 
This 3D full attention approach ensures that all pixels in a sequence of feature maps contribute to the attention map, enabling smooth and coherent motion even in the presence of vigorous movement.

Our approach diverges from some concurrent works~\cite{zheng2024vitonditlearninginthewildvideo} by unifying the image and video VTON within a single framework, eliminating the need for an additional network.

\begin{figure*}[t]
	\centering
	\includegraphics[width=0.95\linewidth]{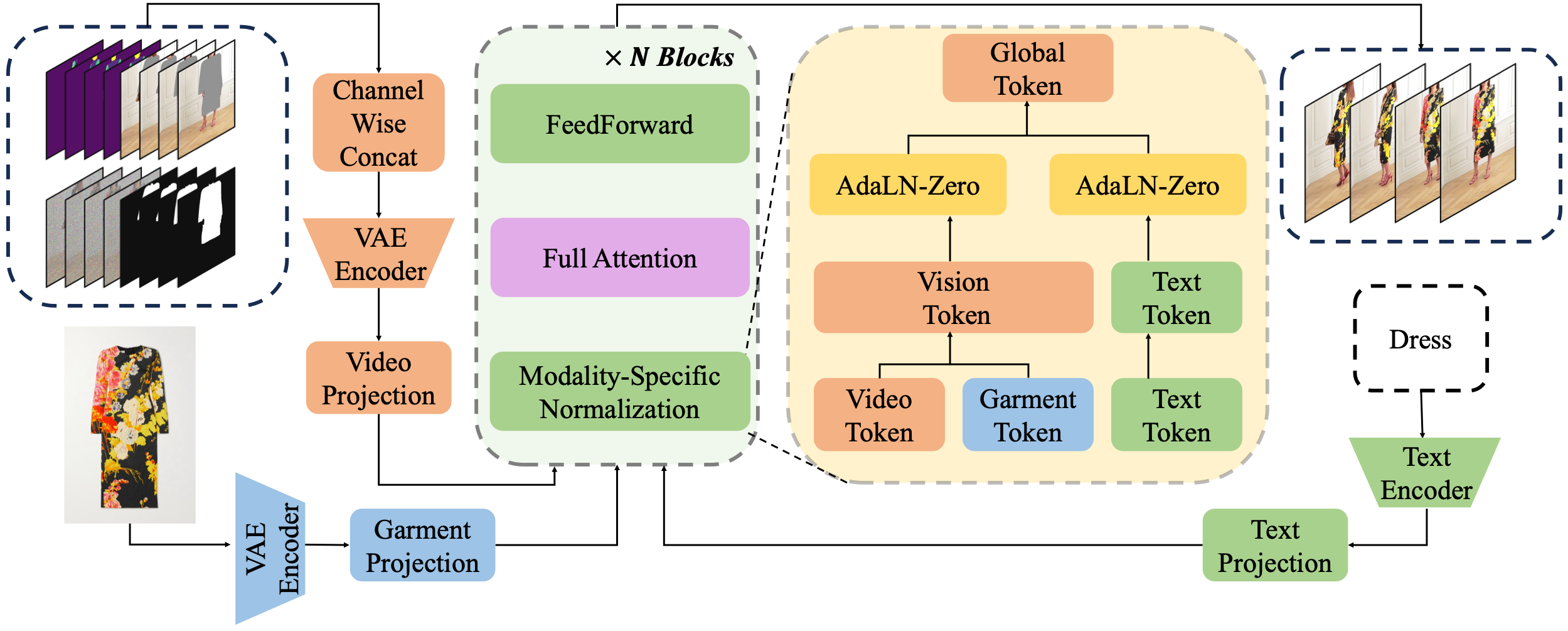}
    \vspace{-2mm}
    \caption{Overview of the proposed MN-VTON. Our method achieves high-quality image and video virtual try-on (VTON) through a Modality-Specific Normalization module. Specifically, for multi-modal inputs, we first apply identical AdaLN-zero normalization for similar modality inputs (\eg, reference garment and image/video) and distinct AdaLN-zero normalization for different modalities (\eg, text and visual inputs). Next, we employ shared-weight self-attention across all tokens to enable effective VTON using only a {\it single} network.}
	\label{fig:pipeline}
    \vspace{-5mm}
\end{figure*}


\section{Method}

As discussed in Sec.~\ref{sec:intro}, state-of-the-art methods address the three limitations {\bf[L1]}, {\bf[L2]}, {\bf[L3]} of early single-network methods with a novel {\bf dual} network paradigm.
In this work, we challenge this paradigm by showing that i) {\bf[L1]} and {\bf[L2]} can be effectively mitigated through a novel layer-wise feature split, normalization, and fusion strategy (Sec.~\ref{sec:token_concate}); ii) {\bf[L3]} can be addressed by a novel modality-specific normalization technique (Sec.~\ref{sec: modality specific}).
Both techniques are implemented within a single network.


\subsection{Preliminaries}
\label{sec:background}
Given a person image/video $I$ and a reference garment $G$, our goal is to generate a new image/video $\hat{I}$ that preserves the original pose and identity of the person while seamlessly replacing their outfit with the reference garment.

\vspace{1mm}
\noindent\textbf{Single-Network Paradigm.} As discussed earlier, single-network methods typically rely on a warping module to adjust the clothing to fit the human body, which can be formulated as follows:
\begin{align}
\begin{split}
    &\hat{I} = \mathrm{VTON}(I_m, M, P, G_w, C_{text}, C_{garment})
\end{split}
\end{align}
where $I_m$ represents the clothing-agnostic version of the input person image/video $I$; $M$ denotes the garment inpainting mask; $P$ denotes the human pose; $G_w$ is the warped garment, typically generated by a garment warping module; $C_{text}$ and $C_{garment}$ are layer-wise CLIP features extracted from the input descriptive text and garment image $G$, respectively.
Note that conventional methods do not use $C_{text}$ and $C_{garment}$ and primarily focus on blending the warped garment onto the person.
Nevertheless, recent diffusion-based approaches, such as LaDI-VTON \cite{morelli2023ladi}, show that incorporating $C_{text}$ and $C_{garment}$ mitigates the feature ``wash-away'' effect during network propagation~\cite{park2019semantic} and is beneficial for VTON.

\vspace{1mm}
\noindent\textbf{Dual-Network Paradigm.} 
To further refine the details of the clothing, an additional {\it ReferenceNet} is employed to extract well-aligned, multi-scale garment features. These features are then integrated with the person's features using a self-attention mechanism, which can be formulated as:
\begin{align}
\begin{split}
    & \mathbf{F^{\rm ref}} = \{F^{\rm ref}_{0}, F^{\rm ref}_{1}, \dots, F^{\rm ref}_{N}\}  = {\rm ReferenceNet}(G) \\
    &\hat{I} = \mathrm{VTON}(I_m, M, P, C_{text}, \mathbf{F^{\rm ref}}) 
\end{split}
\label{eq:dual_network}
\end{align}
where $\mathbf{F} = \{F_{0}, F_{1}, \dots, F_{N}\}$ is the layer-wise features extracted by {\it ReferenceNet}, $N$ is the number of layers.

\vspace{1mm}
\noindent\textbf{Remark.} 
It is important to highlight that pose plays a critical and irreplaceable role in VTON. Specifically, like the third line of Fig. \ref{fig: VITONHD_Compare}, when a person's upper body is occluded or masked, pose information is essential for accurately reconstructing the body and ensuring a realistic garment fit.

\subsection{Feature Split, Normalization and Fusion}
\label{sec:token_concate}

As Eq.~\ref{eq:dual_network} shows, dual-network methods address the limitations of {\bf[L1]} and {\bf[L2]} by introducing an additional {\it ReferenceNet}. This network not only extracts multi-scale garment features but also seamlessly fuses them with those from the {\it Main Network} for further processing.
Let $\mathbf{F^{\rm main}} = \{F^{\rm main}_0, F^{\rm main}_1, \dots, F^{\rm main}_N\}$ be the features of the Main Network, dual-network methods proposed that:
\vspace{2mm}
\begin{align}
\begin{split}
    & Q^l = W_Q^lF^{\rm main}_l\\  
    & K^l = W_K^l[F^{\rm main}_l \oplus F^{\rm ref}_l]\\
    & V^l = W_V^l[F^{\rm main}_l \oplus F^{\rm ref}_l] \\
    & F^{\rm main}_{l+1} = {\rm Attention}(Q^l, K^l, V^l)  
\end{split}
\label{eq:dual_network_attention}
\end{align}
where $\oplus$ denotes feature concatenation along the channel axis; $l$ is the layer ID.

However, the feature extraction and fusion strategy in Eq.~\ref{eq:dual_network_attention} presents a challenge for single-network methods, as their network layers output only a single feature $F^{\rm main}_l$.
To address this issue, a straightforward approach is to introduce a {\it split-and-fusion} strategy, which splits $F^{\rm main}_l$ into two parts: $F^{\rm main,1}_l$ and $F^{\rm main,2}_l$ ($F^{\rm main}_l = F^{\rm main,1}_l \oplus F^{\rm main,2}_l$), and processes them similarly to $F^{\rm main}_l$ and $F^{\rm ref}_l$ in Eq.~\ref{eq:dual_network_attention}, respectively:
\begin{align}
\begin{split}
    & Q^l = W_Q^lF^{\rm main,1}_l\\  
    & K^l = W_K^l[F^{\rm main,1}_l \oplus F^{\rm main,2}_l] \\
    & V^l = W_V^l[F^{\rm main,1}_l \oplus F^{\rm main,2}_l] \\
    & F^{\rm main}_{l+1} = {\rm Attention}(Q^l, K^l, V^l)  
\end{split}
\label{eq:single_network_attention_naive}
\end{align}
Nevertheless, this approach has two major problems: i) {\it feature shrinking}: the dimension of $Q^l$/$F^{\rm main,1}_l$ is smaller than $F^{\rm main}_l$, resulting in the dimension of $F^{\rm main}_{l+1} < F^{\rm main}_l$; ii) since $F^{\rm main,1}_l \oplus F^{\rm main,2}_l = F^{\rm main}_l$, Eq.~\ref{eq:single_network_attention_naive} is not effective as it is essentially a weaker version of the native single-attention mechanism where $Q^l = W_Q^l[F^{\rm main,1}_l \oplus F^{\rm main,2}_l]$.
To alleviate these problems, we have modified Eq.~\ref{eq:single_network_attention_naive} by introducing separate normalization layers for $F^{\rm main,1}_l$ and $F^{\rm main,2}_l$, respectively, and use both of them in $Q$, and have:
\begin{align}
\begin{split}
    & F'_l = {\rm Norm(F^{\rm main,1}_l)} \oplus {\rm Norm(F^{\rm main,2}_l)}\\
    & Q^l = W_Q^lF'_l, K^l = W_K^lF'_l , V^l = W_V^lF'_l \\
    & F^{\rm main}_{l+1} = {\rm Attention}(Q^l, K^l, V^l)  
\end{split}
\label{eq:single_network_attention_improved}
\end{align}
where $F'_l \neq F^{\rm main}_l$ as $F^{\rm main,1}_l$ and $F^{\rm main,2}_l$ are modulated independently based on their distinct characteristics, thereby effectively capturing the advantages of Eq.~\ref{eq:dual_network_attention}.
However, the effectiveness of Eq.~\ref{eq:single_network_attention_improved} depends on the splitting strategy of $F^{\rm main,1}_l$ and $F^{\rm main,2}_l$, which will be discussed in-depth in the following subsection.

\subsection{Modality-Specific Normalization}
\label{sec: modality specific}
\begin{figure}[t]
	\centering
	\includegraphics[width=0.95\linewidth]{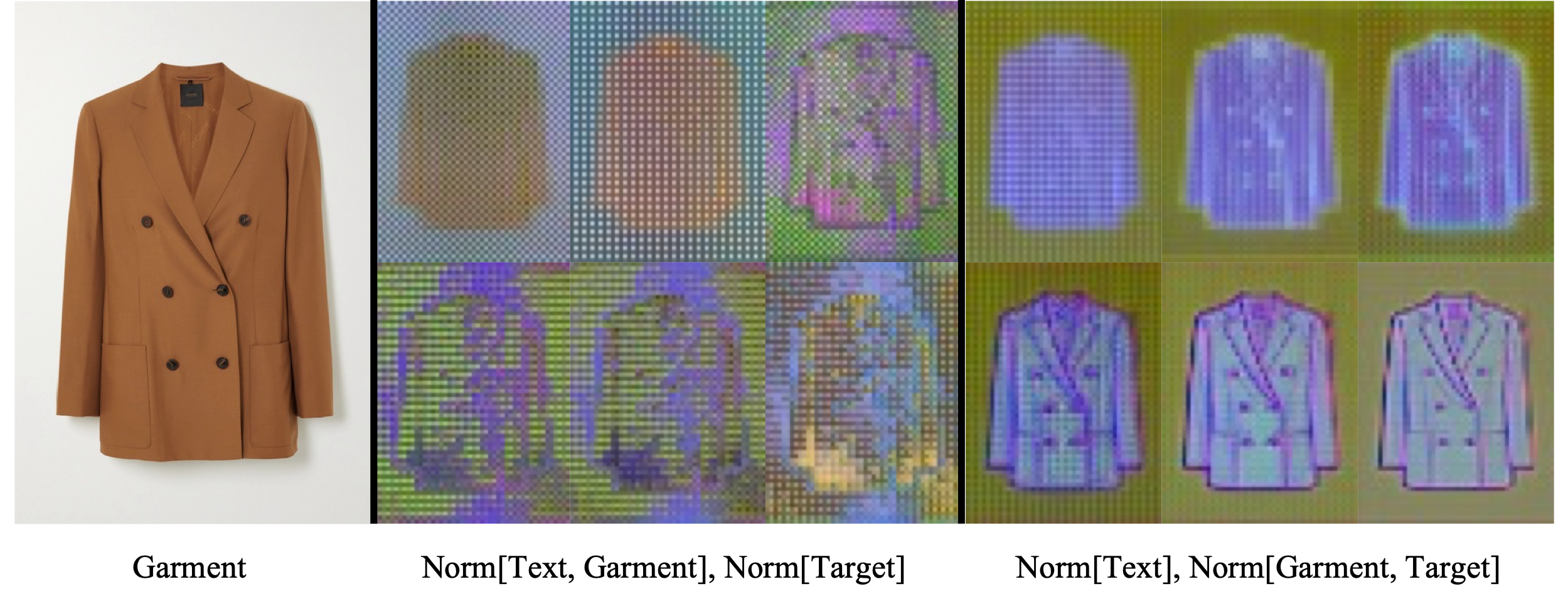}
    	\caption{Visualization of garment feature maps $F^{\rm garment}_l$ using PCA at the output of blocks 1, 6 11, 16, 21, 26 of our MN-VTON. [ $\cdot$ , $\cdot$ ] denotes different combinations of input modalities.}
	\label{fig: feature_visual}
    \vspace{-0.1cm}
\end{figure}

As mentioned above, the splitting strategy of $F^{\rm main,1}_l$ and $F^{\rm main,2}_l$ plays a key role in the success of our method.
Similar to previous works~\cite{xu2024ootdiffusion}, our method also takes three inputs: i) {\it Text}, which describes the type of garment (\eg, upper-body, lower-body, dress); ii) {\it Garment Image}, which is the garment to be worn on the target person; iii) {\it Target Image/Video}, which contains the target person who will wear the garment.
Therefore, we propose to split $F^{\rm main,1}_l$ and $F^{\rm main,2}_l$ according to their modalities and apply normalization to them separately, namely Modality-Specific Normalization, which addresses the limitation {\bf [L3]}.
Specifically, as Fig.~\ref{fig:pipeline} (d) shows, we propose that:
\begin{align}
    &F^{\rm main,1}_l = F^{\rm text}_l \\
    &F^{\rm main,2}_l = F^{\rm garment}_l \oplus F^{\rm target}_l
\end{align}
where $F^{\rm text}_l$, $F^{\rm garment}_l$, and $F^{\rm target}_l$ denote the respective feature components of $F^{\rm main}_l$ corresponding to the text, garment image, and target image or video modalities, respectively.
The rationale for this is from several key insights:
\begin{itemize}
    \item Normalization preserves the relative relationships between different modalities in a multimodal distribution.
    \item Retaining garment and target image or video information throughout the network is essential to capture fine-grained details.
    \item Text information provides complementary semantic context for VTON task.
\end{itemize}
Thus, concatenating $F^{\rm garment}_l$ and $F^{\rm target}_l$ while processing $F^{\rm text}_l$ separately is the most effective approach.
To support this, we visualize the garment feature maps $F^{\rm garment}_l$ using Principal Component Analysis (PCA) for two different splitting strategies in Fig. \ref{fig: feature_visual}. It shows that combining mismatched modalities, such as $F^{\rm text}_l$ and $F^{\rm garment}_l$, introduces blurriness and artifacts in the garment features, whereas aligning similar modalities, like $F^{\rm garment}_l$ and $F^{\rm target}_l$, produces garment features that are both clear and informative.


\vspace{1mm}
\noindent \textbf{Remark.} Another straightforward approach would be to split and normalize each of the three modalities independently. However, as discussed earlier, the Garment Image and Target Image/Video share similar characteristics, and separating them does not yield additional benefits. Please see the supplementary materials for more details.

\begin{table}
  \centering
  
\setlength\tabcolsep{3pt}
\begin{tabular}{|l|r|r|r|}
\hline Methods &  SSIM $\uparrow$ & LPIPS $\downarrow$  & Trainable Param. \\
\hline 
Reference-UNet  & 0.8756  & 5.412  & 1700.25M \\
Ours (UNet)  & 0.8768  & 5.357& 859.57M  \\
Ours (DiT) & 0.8879  & 0.0632  & 1694.28M  \\

\hline
\end{tabular}

\caption{Generalization ability of our method across architectures.}
  \label{tab:network-comparison}
  \vspace{-5mm}
\end{table}

\begin{figure*}[t]
	\centering
	\includegraphics[width=0.95\linewidth]{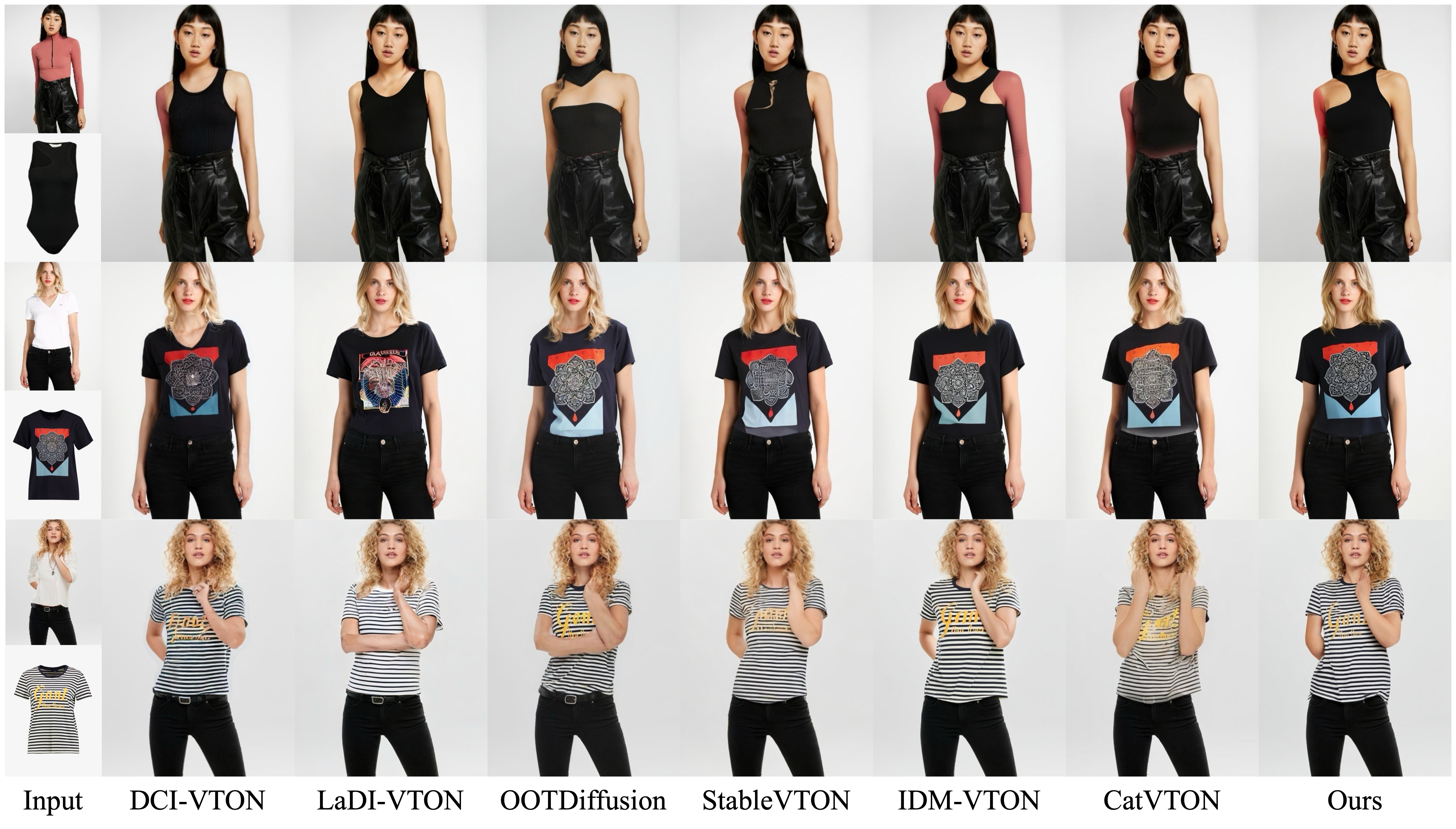}
	\caption{Visual comparison on the VITONHD dataset. Please zoom in for more details.}
	\label{fig: VITONHD_Compare}
\end{figure*}

\subsection{Image-Video Joint training Strategy}
We trained our VTON model on two image datasets and one video dataset (VIVID \cite{fang2024vivid}) for high quality video generation. The image datasets provide a diverse set of cloth-person identities, enhancing generalization, while the video dataset enables the model to learn temporal coherence.
Specifically, we applied position embedding interpolation~\cite{chen2023pixart} to align position encodings between image and video inputs, allowing both to share a common position encoding. This approach improves the model’s ability to generate fine clothing details while preserving temporal consistency across frames.

\begin{figure*}[t]
	\centering
	\includegraphics[width=0.95\linewidth]{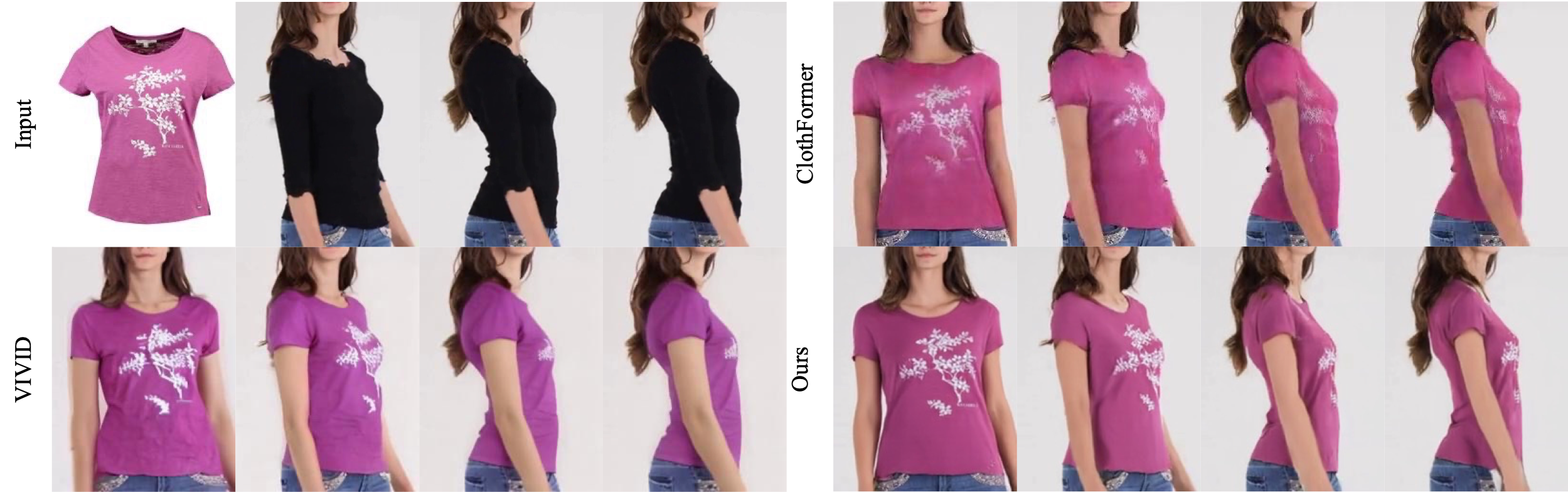}
	\caption{Visual Comparison on VVT dataset.}
	\label{fig:VVT comparison}
    \vspace{-5mm}
\end{figure*}

\begin{figure}[t]
	\centering
	\includegraphics[width=0.95\linewidth]{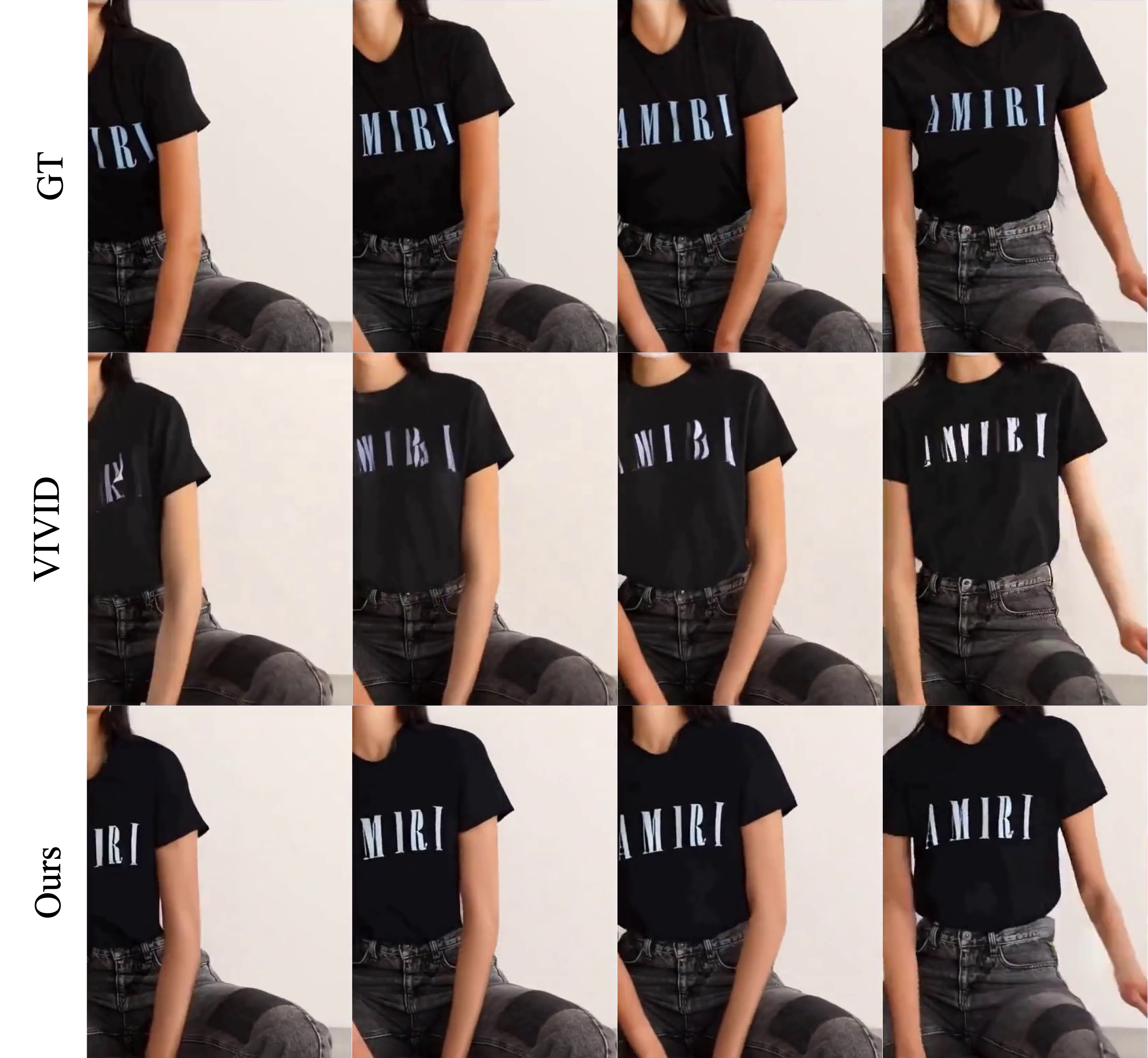}
	\caption{Qualitative comparison on the VIVID dataset.}
	\label{fig:VIVID_comparison}
\end{figure}

\begin{figure}[t]
	\centering
	\includegraphics[width=0.95\linewidth]{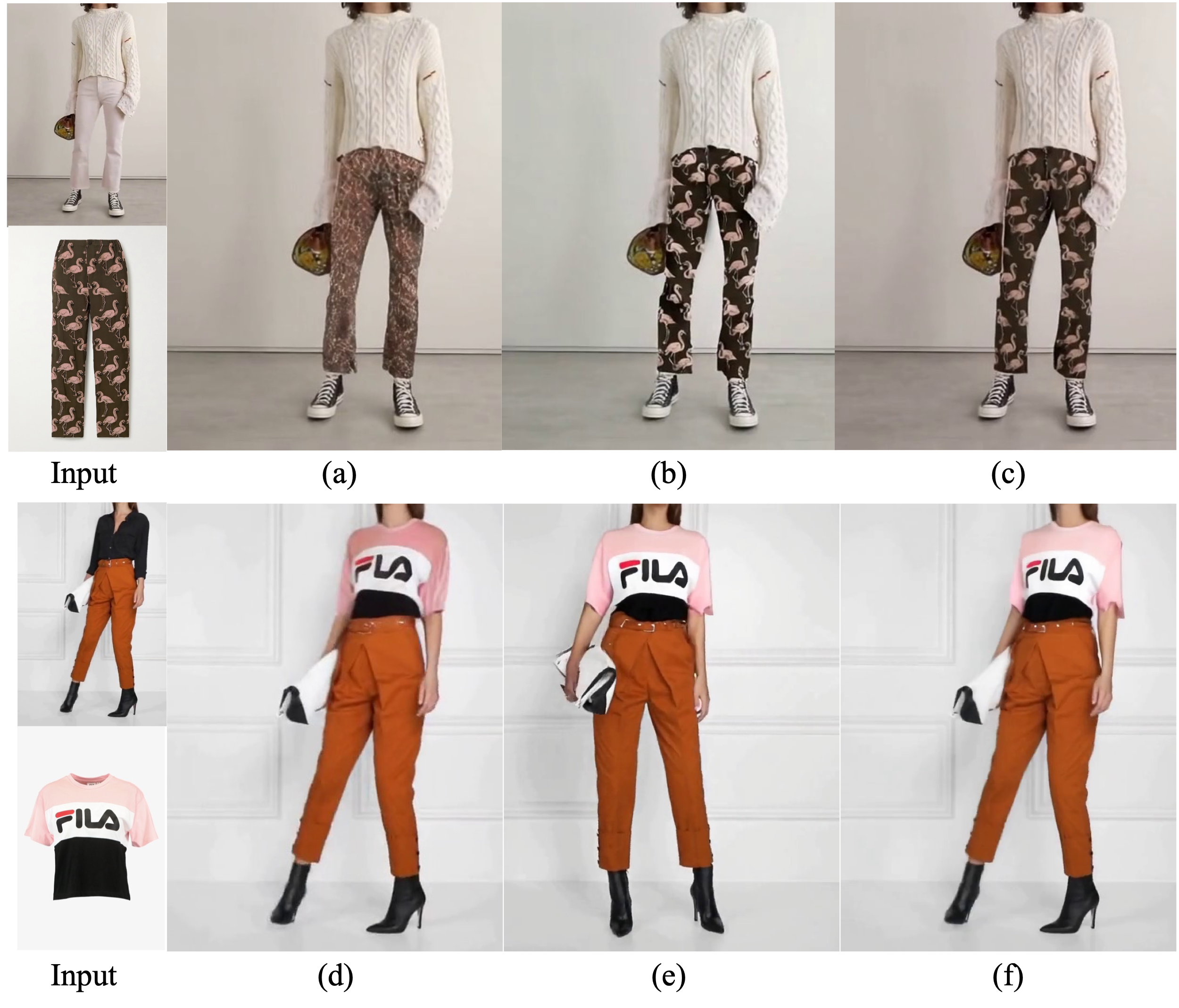}
	\caption{Ablation study on the VIVID dataset. Top row: ``lower body'' as text input; bottom row: ``upper body'' as text input. }
	\label{fig:ablation study}
    \vspace{-5mm}
\end{figure}





\begin{table*}
  \centering
  
\setlength\tabcolsep{3pt}
\begin{tabular}{|l|rrrr|rr|rrrr|rr|}
\hline  & \multicolumn{6}{|c|}{ VTIONHD } & \multicolumn{6}{c|}{DressCode } \\
\hline  & \multicolumn{4}{|c|}{ Paired }  & \multicolumn{2}{|c|}{ Unpaired } & \multicolumn{4}{|c|}{ Paired }  & \multicolumn{2}{|c|}{ Unpaired }  \\
\hline  Method & SSIM $\uparrow$    & FID $\downarrow$  & KID $\downarrow$  & LPIPS $\downarrow$& FID $\downarrow$ & KID $\downarrow$ &  SSIM $\uparrow$    & FID $\downarrow$  & KID $\downarrow$ & LPIPS $\downarrow$ & FID $\downarrow$ & KID $\downarrow$  \\
\hline
DCI-VTON\cite{gou2023taming}   & 0.8620  & 9.408  & 4.547 & 0.0606  & - & -       & -      & -    & - & -      & -  &- \\
StableVTON\cite{kim2024stableviton} & 0.8543  & 6.439  & 0.942 & 0.0905  & - & -       & -      & -    & - & -      & -  &- \\
LaDI-VTON \cite{morelli2023ladi}   & 0.8603  & 11.386  & 7.248 & 0.0733  & 14.648  & 8.754 & 0.7656 & 9.555 & 4.683 & 0.2366  &10.676 & 5.787 \\

IDM-VTON \cite{choi2024improving}  & 0.8499  & 5.762  & 0.732 & 0.0603  & 9.842  & 1.123 & 0.8797 & 6.821 & 2.924 & 0.0563  & 9.546 & 4.320 \\
OOTDiffusion \cite{xu2024ootdiffusion}  & 0.8187  & 9.305  & 4.086 & 0.0876  & 12.408 & 4.689 & 0.8854 & 4.610 & 0.955 & 0.0533  & 12.567 & 6.627 \\
CatVTON\cite{chong2024catvtonconcatenationneedvirtual} & 0.8704 & 5.425  & 0.411 & 0.0565  & 9.015 & 1.091 & 0.8922 & 3.992 & 0.818 & 0.0455 & 6.137 & 1.403 \\
\hline
Ours & 0.8853 & 5.236  & 0.401 & 0.0477  & 8.627 & 0.747 
     & 0.9237 & 2.862 & 0.290 & 0.0291 & 4.966 & 0.961 \\ 
\hline

\end{tabular}
\caption{ Quantitative comparisons for image Virtual Try-On.}
  \label{tab: Image Comparison}
\end{table*}

\begin{table}
  \centering
  
\setlength\tabcolsep{2.4pt}
\begin{tabular}{|l|r|r|r|r|}
\hline Methods &  SSIM $\uparrow$ & LPIPS $\downarrow$ & $FVD_I$ $\downarrow$ &$FVD_R$ $\downarrow$  \\
\hline 
FW-GAN \cite{wu2022fw} & 0.675  & 0.283 & 8.019 & 12.150 \\
ClothFormer \cite{jiang2022clothformer} &  0.921  & 0.081 & 3.967 & 5.048 \\
Tunnel Try-on \cite{xu2024tunnel} &  0.913  & 0.054 & 3.345 & 4.614 \\
VIVID \cite{fang2024vivid} &  0.949  & 0.068 & 3.405 & 5.074 \\
VITON-DiT \cite{zheng2024viton} &  0.896  & 0.080 & 2.498 & 0.187 \\
\bf Ours & \textbf{0.971} & \textbf{0.019} & \textbf{1.926} & \textbf{0.035} \\
\hline
VIVID \cite{fang2024vivid} & 0.8747  & 0.0818 & 1.1394 & 0.0484 \\
\bf Ours & \textbf{0.8879} & \textbf{0.0632} & \textbf{0.9237} & \textbf{0.0362} \\
\hline
\end{tabular}
\caption{Quantitative comparison on the VVT and VIVID datasets.}
  \label{tab:VVT Comparison}
\vspace{-5mm}
\end{table}

\section{Experiment}
\subsection{Experimental Setup}

\vspace{2mm}
\noindent \textbf{Datasets $\&$ Implementation Details.} 
Two image datasets (VITONHD~\cite{choi2021viton}, DressCode~\cite{morelli2022dresscode}) and two video datasets (VVT~\cite{wu2022fw}, VIVID~\cite{fang2024vivid}) are used for evaluation.
Our model is trained based on Cogvideo-2B \cite{yang2024cogvideox}. 
For image VTON task, in order to ensure a fair comparison, we independently trained two models specifically for the VITONHD and DressCode datasets. 
Each model was then evaluated on its respective test set, consistent with the approach adopted by previous methods.
The training and testing resolution are $1,024 \times 768$, and then all images are resized two $512 \times 384$ for metric evaluation. 
All image related experiments were conducted on 8 Nvidia A100 GPUs. We set the learning rate to 2e-4, utilized the AdamW optimizer, and trained with a batch size of 64 for 16,000 steps.
For video VTON task, a clip of 25 frames is used as the video input. 
The training resolution is $512 \times 384$ for VIVID \cite{fang2024vivid} dataset and $256 \times 192$ for VVT \cite{wu2022fw} dataset. 
All video models use hyper-parameters similar to the image models, but the batch size and training steps are adjusted to 16 and 20,000 respectively. 


\subsection{Generalization across Network Architectures} 
\label{model generalization}

To validate the generalization of our method across network architectures, we conducted experiments on the following three methods: i) original ReferenceNet-based diffusion UNet. ii) diffusion Unet with our Modality-specific Normalization (ReferenceNet-free); iii) DiT with our Modality-specific Normalization (ReferenceNet-free ). 
As shown in Table \ref{tab:network-comparison}, by eliminating ReferenceNet and concatenating the features of the garment and target images along the token dimension, 
Ours (UNet) can achieve results comparable to those of ReferenceNet while significantly reducing the trainable parameters; while Ours (DiT) uses a similar number of trainable parameters bus significantly outperforms ReferenceNet. 
Without loss of generality, we chose DiT for our MN-VTON from its superior performance, and use it for all the following experiments.


  



\subsection{Comparison with SOTA Methods}


For image VTON, we compare with diffusion-based methods, (\eg DCI-VTON \cite{gou2023taming}, LaDI-VTON \cite{morelli2023ladi}, IDM-VTON \cite{choi2024improving}, StableVTON \cite{kim2024stableviton}, OOTDiffusion \cite{xu2024ootdiffusion} and CatVTON \cite{chong2024catvtonconcatenationneedvirtual}) as they significantly outperform previous GAN-based methods. 
For video VTON, we collected several state-of-the-art methods, covering GAN-based methods like FW-GAN \cite{wu2022fw}, ClothFormer \cite{jiang2022clothformer} and diffusion-based methods like Tunnel Try-on \cite{xu2024tunnel}, VIVID \cite{fang2024vivid} and VITON-DiT \cite{zheng2024vitonditlearninginthewildvideo}. 

\vspace{1mm}
\noindent \textbf{Image VTON Comparison}
In Fig. \ref{fig: VITONHD_Compare} and Table. \ref{tab: Image Comparison}, we provide qualitative and quantitative comparisons, respectively.
Quantitatively, our methods significantly surpass all SOTA methods by a substantial margin. Qualitatively, our method far surpasses others in maintaining clothing styles, preserving logos, and enriching texture details. For example, in the second row of Fig. \ref{fig: VITONHD_Compare}, it is evident that the logos produced by alternative models often exhibit inaccuracies in color distribution or omit specific details.
In contrast, our method demonstrates superior capability in preserving the intricate details of the clothing.

\vspace{1mm}
\noindent \textbf{Video VTON Comparison}
As illustrated in Table \ref{tab:VVT Comparison} , our method demonstrates substantial advancements over other approaches with respect to the SSIM, LPIPS, and VFID metrics. These improvements underscore the superiority of our approach in maintaining clothing details and ensuring temporal continuity.
Figures \ref{fig:VVT comparison} and \ref{fig:VIVID_comparison} present the qualitative results of various methods applied to the VVT and VIVID datasets, respectively.
 We can see that earlier video VTON techniques often result in blurred outputs with notable color inconsistencies in the generated clothing. Our method, by contrast, distinctly surpasses these others, excelling in both the preservation of clothing details and ensuring temporal continuity. This is especially evident in its ability to retain text accurately.

\begin{figure}[t]
	\centering
	\includegraphics[width=0.95\linewidth]{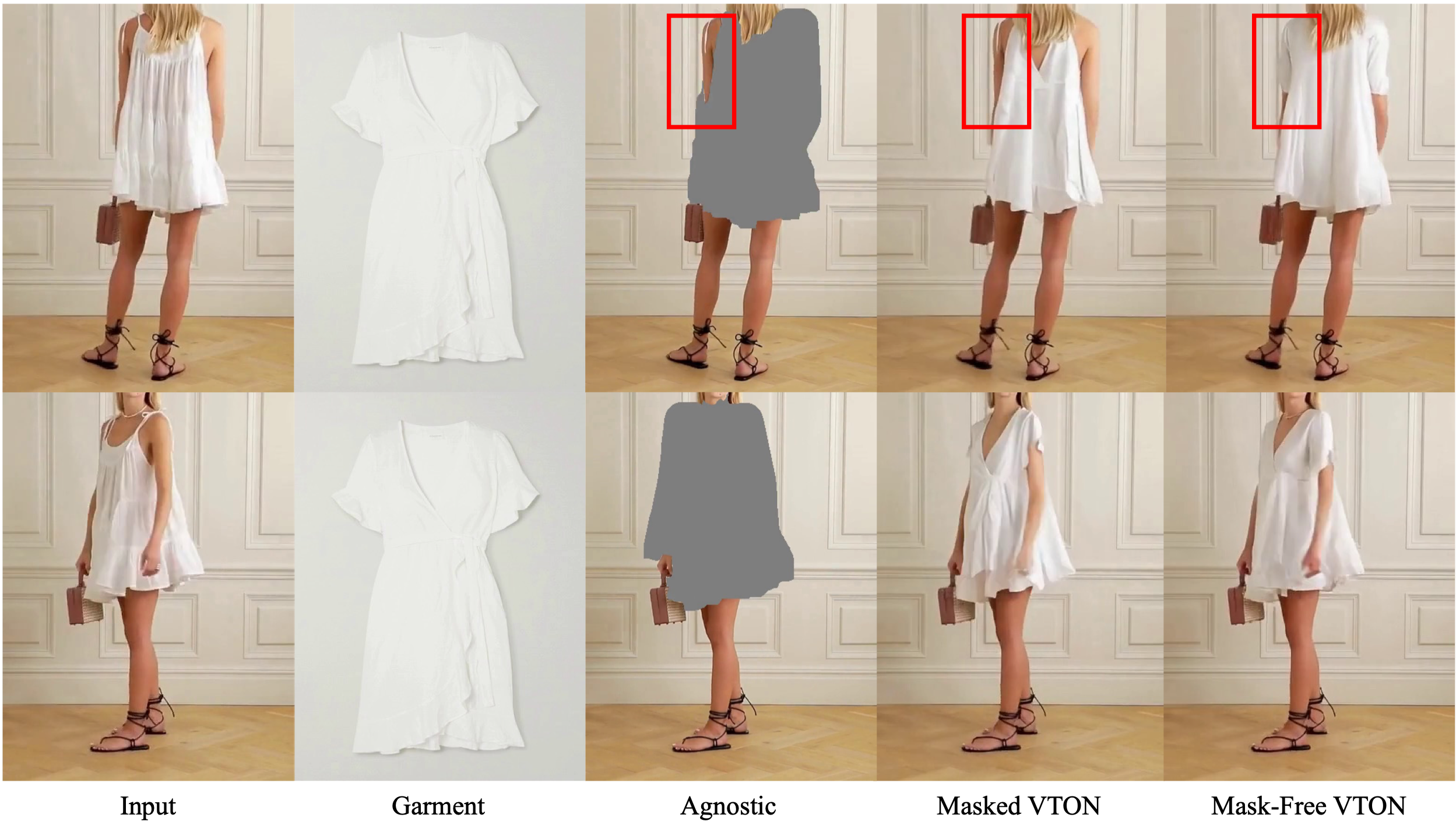}
    
	\caption{Comparison between Masked and Mask-Free VTON.}
	\label{fig:parsing_free_application}
    \vspace{-7mm}
\end{figure}

\subsection{Ablation Study}
We conduct extensive ablations on VIVID \cite{fang2024vivid} dataset to verify the effectiveness of each components, and the visual results is reported in Fig. \ref{fig:ablation study}.

\begin{itemize}
\item (a) and (c):  We explore the integration of text and reference garment as a combined condition for normalization, while the features of video are normalized independently. 
Refer to Fig. \ref{fig: feature_visual}, our findings indicate that normalizing the text and reference image together tends to gradually diminish the garment features, which in turn significantly impacts the quality of the generated results. 
In contrast, by concatenating features of the same modality together and then performing normalization, We can achieve a reasonable result.
\item  (b) and (c):  We observe that image-video joint training yields generated clothing that more closely resembles the reference garments. This outcome can be attributed to the substantial volume of image pair data, which significantly enhances the generalization of our method. 
\item (d) , (e) and (f): (d) and (f) are images from a generated video clip at time $t = 6$, with resolutions of $512 \times 384$ and $832 \times 624$, respectively. (e) is an image from the clip at time $t = 33$.
According to Sec.~\ref{model generalization}, through feature fusion and modality-specific normalization, text, reference garment and video can share the same attention layers, which enable us to generate video clips with higher resolution and longer duration. 
 
\end{itemize}

\begin{table}
  \centering
  
\setlength\tabcolsep{3pt}
\begin{tabular}{|l|r|r|r|}
\hline Methods &  Quality   & Fidelity   & Smoothness   \\
\hline
OOTDiffusion \cite{xu2024ootdiffusion} & 10$\%$  & 7$\%$ & - \\
IDM-VTON \cite{choi2024improving} & 12$\%$  & 14$\%$ & - \\
CatVTON \cite{chong2024catvtonconcatenationneedvirtual} & 13$\%$  & 12$\%$ & - \\
Ours  & 65$\%$  & 67$\%$ & -\\
\hline 
VIVID \cite{fang2024vivid} & 32$\%$  & 25$\%$ & 8$\%$ \\
Ours &  68$\%$  & 75$\%$ &  92$\%$ \\

\hline

\end{tabular}
\caption{User study on preference rates for the VITONHD and VIVID datasets.}
  \label{tab:User Study}
  \vspace{-5mm}
\end{table}



\subsection{User Study}
 As the same setting as Tunnel Try-on \cite{xu2024tunnel}, we conducted two user studies involving 100 participants from diverse backgrounds and age groups to objectively assess our methods against the state-of-the-art techniques in image and video VTON. The evaluation involves three aspects: i) {\it Quality} denotes the image quality; ii) {\it Fidelity} evaluates the ability to preserve garment details. iii) {\it Smoothness} measures the temporal consistency of generated videos. The results indicate that our method considerably outperforms other methods, particularly in terms of video smoothness.

\subsection{Application-Parsing Free Virtual Tryon}
It's quite surprising that our method can generate realistic videos, and one piece of evidence is that our method can be used for mask-free video VTON. We found that one of the biggest challenges for mask-free video VTON is the lack of paired video (\eg two identical videos but with different garments), but our method fills this gap. We can use our well-trained MN-VTON to generate paired data, and then proceed with the training of the parsing-free video VTON.
As shown in Fig. \ref{fig:parsing_free_application}, the effectiveness of masked VTON is highly reliant on the quality of the mask. If there are detection omissions in the mask, achieving accurate generation results becomes challenging. In contrast, mask-free VTON does not suffer from this issue. Additionally, since the mask must fully encompass the clothing area, it often ends up covering regions that do not need alteration(\eg bag,arm). Mask-free VTON effectively avoids this problem.

\section{Conclusion}

In this paper, we proposed a novel single-network approach for virtual try-on (VTON), challenging the need for dual networks to achieve high-quality results. By incorporating a modality-specific normalization strategy, our method efficiently processes and integrates text, image, and video features within a single framework, addressing key challenges in garment realism, texture preservation, and feature alignment.
Extensive experiments demonstrate that our method delivers superior visual quality and detail, matching or surpassing dual-network approaches in both resolution and fidelity, while significantly reducing computational overhead. 
Our work highlights the potential of single-network architectures for scalable, high-performance VTON systems, offering a more efficient solution for real-world applications.

{
    \small
    \bibliographystyle{ieeenat_fullname}
    \bibliography{main}
}


\end{document}


\maketitle

\section{Improved Video Segmentation} 
The video segmentation used in existing video VTON methods suffers from several significant limitations:
i) Garment segmentation, particularly for lower-body clothing, is often inaccurate.
ii) Temporal continuity is substantially compromised due to segmentation being performed at the image level.
iii) Arm segmentation frequently fails when the individual is facing away from the camera.
To address these shortcomings, we propose the following two improvements:
i) For better clothing segmentation, we leverage the integration of GroundingDINO-V1.5 \cite{ren2024grounding} with SAM-2 \cite{ravi2024sam}.
To enhance arm segmentation, we utilize SAPIENS \cite{khirodkar2024sapiens} to generate precise arm masks, followed by post-processing corrections based on the original algorithm.
These modifications collectively enhance the segmentation quality, improving accuracy and continuity across video frames.

\section{Additional Dataset Details}

The VITONHD dataset includes $13,679$ pairs of frontal upper-body images along with their corresponding clothing items, with $2,032$ pairs designated as the test set.
The DressCode dataset contains pairs of full-body images and their corresponding garments, with $15,363$ pairs for upper-body garments, $8,951$ pairs for lower-body garments, and $29,478$ pairs for dresses. 
For each category of garments, $1,800$ pairs are designated as the test set.
The VVT dataset contains 791 paired person videos and garment images at a resolution of 256$\times$192, where 130 videos are used as the test set.
The VIVID dataset consists of $9,700$ paired person videos and garment images at a resolution of 832$\times$624, where $1,941$ videos were allocated as the test set.


\section{Additional Metric Details}
Following VIVID \cite{fang2024vivid}, we employ Structural Similarity Index Measure (SSIM) \cite{wang2004image} and Learned Perceptual Image Patch Similarity (LPIPS) \cite{zhang2018unreasonable} to evaluate image quality. To assess realism and fidelity, we use Kernel Inception Distance (KID) \cite{binkowski2018demystifying} and Fréchet Inception Distance (FID) \cite{heusel2017gans}.
For video results, we adopt Video Fréchet Inception Distance (VFID) \cite{wu2022fw}, which captures both visual quality and temporal consistency, providing a comprehensive evaluation of the generated video content.




\section{Additional Quantitative Results}
To further justify the effectiveness of our Modality-specific Normalization strategy, we conduct a comparison among its three variants as follows:
\begin{itemize}
\item V1: $F'_l = {\rm Norm(F^{\rm text}_l \oplus F^{\rm garment}_l)} \oplus  {\rm Norm(F^{\rm target})}$
\item V2: \\$F'_l = {\rm Norm(F^{\rm text})} \oplus {\rm Norm(F^{\rm garment}_l)} \oplus {\rm Norm(F^{\rm target}_l)}$
\item V3 (ours): \\$F'_l = {\rm Norm(F^{\rm text})} \oplus {\rm Norm(F^{\rm garment}_l \oplus F^{\rm target}_l)}$

\end{itemize}
where $F'_l$ are the combined features (Eq. 5, main paper).

As shown in Table \ref{tab:normalization ablation}, i) V2 and V3 (ours) yield similar results, suggesting that inputs with similar modalities (\ie, image and video) can be grouped together for normalization, thereby improving both training and inference; ii) V1, however, performs significantly worse, highlighting the necessity of the proposed Modality-specific Normalization to effectively handle inputs from different modalities.
We use V3 in our main paper as it performs the best.

\section{Additional Qualitative Results}
Fig.~\ref{fig: DressCode_compare} shows a comparison between our method and other state-of-the-art (SOTA) approaches on the DressCode dataset. It can be observed that our model demonstrates superior performance in maintaining garment types, including accurately estimating clothing length and preserving texture details across various clothing categories (\eg, upper-body, lower-body, and dresses).


\section{Demo}
We include a demo in the supplementary materials (named 1-2-1-demo.mp4) to demonstrate the comparison with SOTA methods.
The demo highlights the effectiveness of our video VTON approach, demonstrating results on long-duration sequences (6 seconds per single inference without video extension) and high-resolution outputs ($832 \times 624$).

\begin{figure*}[t]
	\centering
	\includegraphics[width=0.95\linewidth]{images/VITONHD_Supp.png}
	\caption{Additional qualitative results on the VITONHD dataset.}
	\label{fig:VITON_Supp}
\end{figure*}

\begin{figure*}[t]
	\centering
	\includegraphics[width=0.95\linewidth]{images/DressCode.png}
	\caption{Additional qualitative comparison on the DressCode dataset.}
	\label{fig: DressCode_compare}
\end{figure*}

  

\begin{table}
  \centering
  
\setlength\tabcolsep{2.4pt}
\begin{tabular}{|l|r|r|r|r|}
\hline Methods &  SSIM $\uparrow$ & LPIPS $\downarrow$ & $FVD_I$ $\downarrow$ &$FVD_R$ $\downarrow$  \\
\hline 
V1 & 0.8513  & 0.0986 & 10.7024 & 2.0376 \\
V2  &  0.8865  & 0.0618 & 0.9240 & 0.0358 \\
\bf V3 (ours) & \textbf{0.8879} & \textbf{0.0632} & \textbf{0.9237} & \textbf{0.0362} \\
\hline
\end{tabular}
\caption{Justification of the proposed Modality-specific Normalization strategy.}
  \label{tab:normalization ablation}
\end{table}

{
    \small
    \bibliographystyle{ieeenat_fullname}
    \bibliography{main}
}
